\documentclass[runningheads]{llncs}
\usepackage{graphicx}
\usepackage{pgfplots}
\pgfplotsset{compat=newest}
\usetikzlibrary{matrix}
\usepackage{amsmath,amssymb,amsfonts}
\usepackage{subfig}
\usepackage{latexsym}
\usepackage{url}
\usepackage{algorithm}
\usepackage{algpseudocode}
\usepackage{graphicx}
\usepackage{textcomp}
\usepackage{xcolor}

\usepackage{url}

\usepackage{multirow}
\usepackage{makecell}

\begin{document}
\title{Planning Landmark Based Goal Recognition Revisited: Does Using Initial State Landmarks Make Sense?}%\thanks{Supported by organization x.}}
\titlerunning{Planning Landmark Based Goal Recognition Revisited}
\author{Nils Wilken\inst{1}\orcidID{0000-0003-1336-245X} \and
Lea Cohausz\inst{2} \and
Christian Bartelt\inst{1}\orcidID{0000-0003-0426-6714} \and
Heiner Stuckenschmidt\inst{2}\orcidID{0000-0002-0209-3859}}
\authorrunning{N. Wilken et al.}
\institute{Institute for Enterprise Systems, University of Mannheim, Mannheim, Germany \and
Data and Web Science Group, University of Mannheim, Mannheim, Germany
\email{\{firstname.lastname\}@uni-mannheim.de}}
\maketitle              % typeset the header of the contribution
\begin{abstract}
Goal recognition is an important problem in many application domains (e.g., pervasive computing, intrusion detection, computer games, etc.).
In many application scenarios, it is important that goal recognition algorithms can recognize goals of an observed agent as fast as possible.
However, many early approaches in the area of Plan Recognition As Planning, require quite large amounts of computation time to calculate a solution.
Mainly to address this issue, recently, Pereira et al. \cite{pereira2020landmark} developed an approach that is based on planning landmarks and is much more computationally efficient than previous approaches.
However, the approach, as proposed by Pereira et al., also uses trivial landmarks (i.e., facts that are part of the initial state and goal description are landmarks by definition).
In this paper, we show that it does not provide any benefit to use landmarks that are part of the initial state in a planning landmark based goal recognition approach.
The empirical results show that omitting initial state landmarks for goal recognition improves goal recognition performance.

\end{abstract}
%
%
%

%\newpage
\section{Introduction}
\label{sec:introduction}
Goal recognition is the task of recognizing the goal(s) of an observed agent from a possibly incomplete sequence of actions executed by an observed agent.
This task is relevant in many real-world application domains like crime detection \cite{geib2001plan}, pervasive computing \cite{wilken2021hybrid},\cite{geib2002problems}, or traffic monitoring \cite{pynadath1995accounting}.
State-of-the-art goal recognition systems often rely on the principle of Plan Recognition As Planning (PRAP) and hence, utilize classical planning systems to solve the goal recognition problem \cite{ramirez2009plan}, \cite{ramirez2010probabilistic}, \cite{sohrabi2016revisited}, \cite{amado2018lstm}.
However, many of these approaches require quite large amounts of computation time to calculate a solution.
Mainly to address this issue, recently, Pereira et al. \cite{pereira2020landmark} developed an approach that is based on planning landmarks (PLR) and is much more computationally efficient than previous approaches.
The approach, as proposed by Pereira et al., also uses trivial landmarks (i.e., facts that are part of the initial state and goal description are landmarks by definition).
However, in this paper, we formally analyze and discuss why it does not provide any benefit using initial state landmarks for goal recognition.
On the contrary, we show that ignoring initial state landmarks is superior regarding goal recognition performance.
In addition, we provide three new evaluation datasets and analyze how the structure of a goal recognition problem affects the results of a planning landmark based goal recognition approach when initial state landmarks are used or ignored.
More explicitly, the contributions of this paper are:
\begin{enumerate}
    \item We formally discuss why it does not provide a benefit to use initial state landmarks for goal recognition and propose and adjusted planning landmark based approach.
    \item We provide three new benchmark datasets that are based on a publicly available dataset, which is commonly used in the literature \cite{zenodo}.
    These datasets have a modified goal structure, such that not all possible goals include the same number of facts, which has an effect onto evaluation performance.
    \item We empirically show that ignoring initial state landmarks is superior regarding goal recognition performance of the PLR approach.
\end{enumerate}

\section{Background}
\label{sec:background}
In the context of classical planning systems, planning landmarks are usually utilized to guide the heuristic search through the search space that is induced by a planning problem \cite{hoffmann2004ordered}.
However, recently, Pereira et al. \cite{pereira2020landmark} proposed an approach that utilizes them to solve the goal recognition problem.
The basic idea of PLR is to use the structural information that can be derived from planning landmarks, which can be - informally - seen as way-points that have to be passed by every path to a possible goal.
Hence, when it was observed that such way-points were passed recently by the observed agent, this indicates that the agent currently follows a path to the goal(s) for which the observed way-point is a landmark.
In this work, we propose an adapted version of PLR \cite{pereira2020landmark}.
%Originally, this approach was developed for the goal recognition problem.
%However, in this work, we show that it is easily applicable to the online goal recognition problem, which we consider in the empirical evaluation.
Although PLR was originally developed for the goal recognition problem, it can also be applied to the online goal recognition problem, which we consider in the empirical evaluation.
Before we formally define the goal recognition problem and online goal recognition problem, we start by defining a planning problem.

\subsection{Classical Planning}
\label{subsec:symbolicPlanning}
Classical planning is usually based on a model of the planning domain that defines possible actions, their preconditions, and effects on the domain.
More formally, in this work, we define a (STRIPS) planning problem as follows:
\begin{definition}[(STRIPS) Planning Problem]
A Planning Problem is a tuple $P = \langle F, s_0, A, g \rangle$ where $F$ is a set of facts, $s_0 \subseteq F$ and g $\subseteq F$ are the initial state and a goal, and $A$ is a set of actions with preconditions $Pre(a) \subseteq F$ and lists of facts $Add(a) \subseteq F$ and $Del(a) \subseteq F$ that describe the effect of action $a$ in terms of facts that are added and deleted from the current state.
Actions have a non-negative cost $c(a)$.
A state is a subset of $F$.
A goal state is a state $s$ with $s \supseteq g$.
An action $a$ is applicable in a state $s$ if and only if $Pre(a) \subseteq s$.
Applying an action $a$ in a state $s$ leads to a new state $s' = (s \cup Add(a) \setminus Del(a))$.
A solution for a planning problem (i.e., a plan) is a sequence of applicable actions $\pi = a_1, \cdots a_n$ that transforms the initial state into a goal state.
The cost of a plan is defined as $c(\pi) = \sum \limits_i c(a_i)$.
%A plan is optimal if the cost of the plan is minimal.  
\end{definition}

\subsection{Goal Recognition}
\label{subsec:goalRecognition}
\begin{definition}[Goal Recognition]
\label{def:goalRecognition}
\textit{Goal recognition} is the problem of inferring a nonempty subset $\mathbf{G}$ of a set of intended goals $G$ of an observed agent, given a possibly incomplete sequence of observed actions $\mathbf{O}$ and a domain model $D$ that describes the environment in which the observed agent acts.
Further, the observed agent acts according to a hidden policy $\delta$.
More formally, a goal recognition problem is a tuple $R = \langle D, \mathbf{O}, G \rangle$.
A solution to a goal recognition problem $R$ is a nonempty subset $\mathbf{G} \subseteq G$ such that all $g \in \mathbf{G}$ are considered to be equally most likely to be the true hidden goal $g_*$ that the observed agent currently tries to achieve.
\end{definition}
The most favorable solution to a goal recognition problem $R$ is a subset $\mathbf{G}$ which only contains the true hidden goal $g_*$.
In this work, $D = \langle F, I, A \rangle$ is a planning domain with a set of facts $F$, the initial state $I$, and a set of actions $A$.
The online goal recognition problem is an extension to the previously defined goal recognition problem that additionally introduces the concept of time and we define it as follows:
\begin{definition}[Online Goal Recognition]
\label{def:onlineGoalRecognition}
\textit{Online goal recognition} is a special variant of the \textit{goal recognition} problem (Definition \ref{def:goalRecognition}), where we assume that the observation sequence $\mathbf{O}$ is revealed incrementally.
More explicitly, let $t \in [0,T]$ be a time index, where $T = |\mathbf{O}|$ and hence, $\mathbf{O}$ can be written as $\mathbf{O} = (\mathbf{O}_t)_{t\in [0,T]}$.
For every value of $t$, a goal recognition problem $R(t)$ can be induced as $R(t) = \langle D, G, \mathbf{O_t} \rangle$ where $\mathbf{O_t} = (\mathbf{O}_t)_{t\in [0,t]}$.
A solution to the online goal recognition problem is the nonempty subsets $\mathbf{G}_t \in G; \forall t \in [0,T]$.
\end{definition}

\subsection{Planning Landmarks}
\label{subsec:planningLandmarks}
Planning landmarks are typically defined as facts that must be true (i.e., part of the current planning state) or actions that must be executed at some point during the execution of a valid plan starting at $s_0$ that achieves the goal $g$ \cite{hoffmann2004ordered}.
PLR only focuses on \textit{fact landmarks}.
More precisely, following \cite{hoffmann2004ordered}, we define fact landmarks as follows:
\begin{definition}[Fact Landmark]
\label{def:planningLandmark}
Given a planning problem $P = \langle F, s_0, A, g \rangle$, a fact $f \in F$ is a fact landmark if for all plans $\pi = \langle a_1, \dots, a_n \rangle$ that reach $g$: $\exists s_i: f \in s_i; 0 \leq i \leq n$, where $s_i$ is the planning state that is reached by applying action $a_i$ to state $s_{i-1}$. 
\end{definition}
\cite{hoffmann2004ordered} further divide this set of fact landmarks into \textit{trivial} and \textit{non-trivial} landmarks.
They consider all landmarks that are either contained in the initial state (i.e., $f \in s_0$) or are part of the goal (i.e., $f \in g$) as trivial landmarks because they are trivially given by the planning problem definition.
All other landmarks are considered to be non-trivial.
\begin{figure}[htbp]
    \centering
    \includegraphics[width=0.5\linewidth]{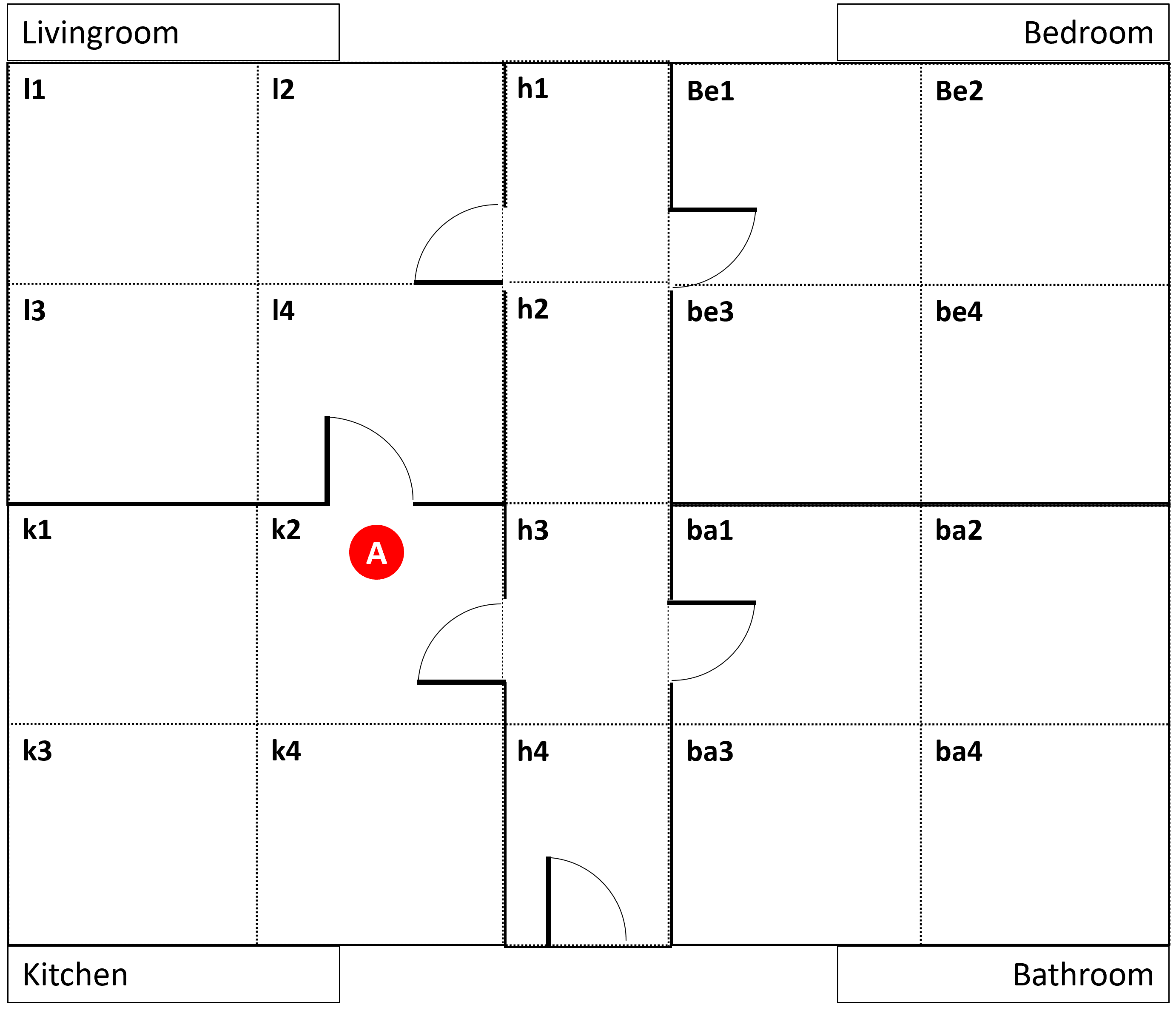}
    \caption{Exemplary Smart Home Layout.}
    \label{fig:LandmarkExample}
\end{figure}
As an example, consider the smart home scenario depicted in Figure \ref{fig:LandmarkExample}.
For this example, we assume, that the corresponding planning domain uses a predicate \textit{(is-at ?x)} to describe the current position of the agent (e.g., in the depicted state the grounded fact \textit{(is-at k2)} is true).
For this example, one potential goal of the agent is defined as $g = \{$\textit{(is-at ba3)}$\}$.
When we assume that the agent can carry out movement actions from one cell to any adjacent cell, then the facts \textit{(is-at h3)} and \textit{(is-at ba1)} would be \textit{non-trivial} fact landmarks because these cells have to be visited by every valid path from the initial position k2 to the goal position ba3 but are not part of the initial state or the goal.
Moreover, \textit{(is-at k2)} and \textit{(is-at ba3)} would be \textit{trivial} landmarks because they also have to be true on every valid path but they are given in the initial state and the goal definition respectively.

\subsection{Extraction of Planning Landmarks}
\label{subsec:extractingPlanningLandmarks}
To extract landmarks, we use two landmark extraction algorithms, which were also used by Gusm{\~a}o et al. \cite{gusmao2020more}.
Both algorithms will be described shortly in the following.

\textit{Exhaustive.}
This algorithm computes an exhaustive set of fact landmarks on the basis of a Relaxed Planning Graph (RPG).
An RPG is a relaxed representation of a planning graph that ignores all delete effects.
As shown by Hoffmann et al. \cite{hoffmann2004ordered}, RPGs can be used to check whether a fact is a landmark.
The exhaustive algorithm checks for every fact $f \in F$, whether it is a landmark.

%\textit{Zhu and Givan \cite{zhu2003landmark}.}
%Zhu and Givan developed a landmark extraction approach that is based on propagating labels along the edges of a planning graph.
%Such labels represent either a fact or an action.
%The idea of this algorithm is to label each action or fact in any action or fact level of the planning graph with the set of all actions and facts that have to occur on any previous level to reach a fact or fulfill the preconditions of an action.
%Hence, the union of the sets of facts by which facts from the goal are labeled, are found to be fact landmarks.

%\textit{Hoffmann et al. \cite{hoffmann2004ordered}.}
%Similarly to the exhaustive algorithm, described above, Hoffmann et al. also propose to extract landmarks on the basis of a RPG.
%However, in contrast to the exhaustive algorithm, the algorithm proposed by Hoffmann et al. does not check for all facts whether they are landmarks.
%Instead, they propose to only check facts that are part of the preconditions of actions that are determined as first achiever actions of any facts that were already added as landmarks.
%Hence, on the one hand, as this algorithm does not check all facts, it is computationally more efficient than the exhaustive algorithm.
%On the other hand, the algorithm only detects a subset of the set of all fact landmarks.

\textit{Richter \cite{richter2008landmarks}.}
The algorithm proposed by Richter in principle works similarly to the algorithm developed by Hoffmann et al. \cite{hoffmann2004ordered}, which was originally used by Pereira et al. \cite{pereira2020landmark}.
The two main differences are that the algorithm by Richter considers the $SAS^+$ encoding of planning domains and allows disjunctive landmarks.
The algorithm of Hoffmann et al. only considers facts as potential landmarks that are part of the preconditions of all first achievers of a potential landmark $l$.
In contrast, the algorithm proposed by Richter allows for disjunctive landmarks, where each disjunctive landmark contains one fact from one precondition of one of the possible achievers of $l$.
This allows this method to find more landmarks than the algorithm from Hoffmann et al.

\section{Ignoring Initial State Landmarks in Planning Landmark Based Goal Recognition}
\label{sec:plrgoalrecognition}
In this paper, we propose to adjust PLR such that initial state landmarks are ignored.
The following subsections first introduce the adjusted approach before Subsection \ref{subsec:usingInitialLandmarks} formally analyzes and discusses, why we think that considering initial state landmarks provides no additional benefit to solve the goal recognition problem.
%The following subsection analyzes and discusses, why we think that considering initial state landmarks provides no additional benefit to solve the goal recognition problem and might even have a negative impact on the recognition performance.

\subsection{Computing Achieved Grounded Landmarks}
\label{subsec:computingAchievedLandmarks}
The two heuristics, which were proposed by Pereira et al. \cite{pereira2020landmark} to estimate $P(G|O)$, both reason over the set of landmarks that were already achieved by a given observation sequence $\pmb{o}$ for each goal $g \in G$, which is referred to as $AL_g$.
To determine the set of achieved landmarks for each goal, we use Algorithm \ref{alg:computeAchievedLandmarks}.
This algorithm is inspired by the original algorithm proposed by Pereira et al. \cite{pereira2020landmark}.
\begin{algorithm}
\caption{Compute achieved landmarks for each goal.}
\label{alg:computeAchievedLandmarks}
\textbf{Input: } \textit{$I$ initial state, $G$ set of candidate goals, $\pmb{o}$ observations, and a set of extracted landmarks $L_g$ for each goal $g \in G$}. \\
\textbf{Output: } \textit{A mapping $M_G$ between each goal $g \in G$ and the respective set of achieved landmarks $AL_g$.}
\begin{algorithmic}[1]
\Function{Compute Achieved Landmarks}{$I$, $G$, $\pmb{o}$, $L_G$}
\State $M_G \leftarrow \langle \rangle$
\ForAll{$g \in G$}
    \State $L_g \leftarrow$ all fact landmarks from $L_g$ s.t.
    \State \hspace*{1cm}$\forall l \in L_g: l \notin I$
    \State $L \leftarrow \emptyset$
    \State $AL_g \leftarrow \emptyset$
    \ForAll{$o \in \pmb{o}$}
        \State $L \leftarrow \{l \in L_g|l \in Pre(o) \cup Add(o) \wedge l \notin L\}$
        \State $AL_g \leftarrow AL_g \cup L$
    \EndFor
    \State $M_G(g) \leftarrow AL_g$
\EndFor
\State \Return $M_G$
\EndFunction
\end{algorithmic}
\end{algorithm}
Compared to the original, Algorithm \ref{alg:computeAchievedLandmarks} differs substantially in two points.
First, it is not able to consider the predecessor landmarks for each landmark that was detected to be achieved by the given observations.
The reason for this is that ordering information between landmarks would be necessary to do this.
However, such information is not generated by all landmark extraction methods that are evaluated in this paper.
As a consequence, the adjusted algorithm will probably have more difficulties dealing with missing observations compared to the original algorithm.
Second, in contrast to the original algorithm, Algorithm \ref{alg:computeAchievedLandmarks} does not consider initial state landmarks to be actually \textit{achieved} by the given observation sequence $\pmb{o}$.
Instead, these landmarks are simply ignored during the goal recognition process.

\subsection{Estimating Goal Probabilities}
\label{subsec:estimatingGoalProbabilities}
To estimate the goal probabilities from the sets of all extracted landmarks (i.e., $L_g$) and landmarks already achieved by $\pmb{o}$ (i.e., $AL_g$) for each $g \in G$, we use slightly adjusted versions of the heuristics introduced by \cite{pereira2020landmark}.
%One heuristic considers the percentage of completion in terms of the fraction of all landmarks that were already identified as achieved by the given observation sequence.
%The second heuristic computes a uniqueness score for each landmark and uses these scores for the computation of the heuristic scores.

\textit{Goal Completion Heuristic.}
The original version of this heuristic estimates the completion of an entire goal as the average of completion percentages of the sub-goals of a goal.% (i.e., all facts $sg \in g$, where $g \in G$) of a goal.
More precisely, the original heuristic is computed as follows \cite{pereira2020landmark}:
\begin{equation}
\label{eq:originalCompletionHeuristic}
    h_{gc}(g, AL_g, L_g) = \Bigg( \frac{\sum_{sg \in g}{\frac{|AL_{sg}|}{|L_{sg}|}}}{|g|} \Bigg)
\end{equation}
However, to which of the sub-goals each of the identified achieved landmarks contributes can again only be determined if ordering information between the landmarks is available.
Hence, not all landmark extraction methods that are used in this work do generate such information, the completion was slightly adjusted to be computed as:
\begin{equation}
\label{eq:adjustedCompletionHeuristic}
    h_{gc}(g, AL_g, L_g) = \Bigg( \frac{|AL_g|}{|L_g|} \Bigg)
\end{equation}
This adjustment, in some cases, has a significant impact on the resulting heuristic scores.
For example, consider the case that $g = \{sg_0, sg_1, sg_2, sg_3, sg_4\}$, $|L_{sg_i}| = 1$ and $|AL_{sg_i}| = 1$, $\forall sg_{i} \in g; 0 \leq i \leq 3$, $|AL_{sg_4}| = 0$, and $|L_{sg_4}| = 30$.
In this case, the result of Equation \ref{eq:originalCompletionHeuristic} would be $4/5$, whereas the result of Equation \ref{eq:adjustedCompletionHeuristic} would be $4/34$.
Thus, the more unevenly the landmarks are distributed over the sub-goals, the larger the difference between the original heuristic calculation and the adjusted calculation becomes.
Nevertheless, it is not fully clear which of the two options achieves better goal recognition performance.

\textit{Landmark Uniqueness Heuristic.}
The second heuristic, which was proposed by Pereira et al. \cite{pereira2020landmark}, does not only consider the percentage of completion of a goal in terms of achieved landmarks but also considers the uniqueness of the landmarks.
The intuition behind this heuristic is that it is quite common that several goals share a common set of fact landmarks.
Hence, landmarks that are only landmarks of a small set of potential goals (i.e., landmarks that are more unique) provide us with more information regarding the most probable goal than landmarks that are landmarks for a larger set of goals.
For this heuristic, \textit{landmark uniqueness} is defined as the inverse frequency of a landmark among the found sets of landmarks for all potential goals.
More formally the landmark uniqueness is computed as follows \cite{pereira2020landmark}:
\begin{equation}
    \label{eq:landmarkUniquenessScore}
    L_{uniq}(l,L_G) = \Bigg(\frac{1}{\sum_{L_g \in L_G}{|\{l|l \in L_g\}|}}\Bigg)
\end{equation}
Following this, the uniqueness heuristic score is computed as:
\begin{equation}
    \label{eq:uniquenessHeuristic}
    h_{uniq}(g, AL_g, L_g, L_G) = \Bigg(\frac{\sum_{al \in AL_g}{L_{uniq}(al, L_G)}}{\sum_{l \in L_g}{L_{uniq}(l, L_G)}}\Bigg)
\end{equation}
To determine the set of most probable goals, for both heuristics, the heuristic values are calculated for all potential goals.
Based on these scores, the set of goals that are assigned with the highest heuristic score are considered as most probable goals.

\subsection{Why Using Initial State Landmarks Does Bias Goal Recognition Performance}
\label{subsec:usingInitialLandmarks}
We propose to adjust the original PLR approach to ignore initial state landmarks because we think that landmarks that are part of the initial state do not provide any valuable information for goal recognition but might potentially even have a misleading effect.
This is because using initial state landmarks for goal recognition in fact means that information which is not derived from the observed agent behaviour is used for recognition.
Moreover, due to how the two recognition heuristics and the utilized planning domain are defined, using initial state landmarks introduces a bias towards considering goals with smaller numbers of non initial state landmarks as more probable.
As a consequence, the goal(s) that have the largest fraction of their landmarks in the initial state are considered to be most probable when no action has been observed so far.
However, this is only due to how the domain and goal descriptions are defined and not by actually observed agent behaviour.

In the following, this issue is analyzed more formally based on the completion heuristic.
As the uniqueness heuristic is very similar to the completion heuristic, just that it weights more unique landmarks stronger, the theoretical analysis would basically follow the same lines.
Nevertheless, as initial state landmarks have the lowest uniqueness score, the uniqueness heuristic already is closer to ignoring initial state landmarks than the completion heuristic.
\begin{equation}
\label{eq:completionReformulated}
h_{gc}(g,al_g,l_g,s_0) = \frac{|al_g| + |s_0|}{|l_g| + |s_0|}
\end{equation}
The completion heuristic (c.f., Equation \ref{eq:adjustedCompletionHeuristic}) can be reformulated as in Equation \ref{eq:completionReformulated}.
Here, we split the two sets $AL_g$ and $L_g$ into the sets $al_g$ and $s_0$, and $l_g$ and $s_0$ respectively.
Consequently, we have $al_g = \{f|f\in AL_g \setminus s_0\}$, $l_g = \{f | f\in L_g \setminus s_0\}$.

Let us now first consider what happens to the heuristic value of the completion heuristic, when we set $|s_0|$ to extreme values (i.e., $0$ or $\infty$) (c.f., equations \ref{eq:cToZero} and \ref{eq:ctoInfinity}).
\begin{equation}
\label{eq:cToZero}
    \lim_{|s_0|\to 0} \frac{|al_g|+|s_0|}{|l_g|+|s_0|} = \frac{|al_g|}{|l_g|}
\end{equation}
When we consider $|s_0|\to 0$, the completion heuristic converges to the value of the fraction $\frac{|al_g|}{|l_g|}$.
This case is similar to ignoring initial state landmarks.% and exactly what we want, when we want to base our decision on which goal is the most probable solely on the information that can be gained from landmarks that were achieved by observed actions.
\begin{equation}
\label{eq:ctoInfinity}
    \lim_{|s_0|\to\infty} \frac{|al_g|+|s_0|}{|l_g|+|s_0|} = 1
\end{equation}
When we consider $|s_0| \to \infty$, the completion heuristic converges to the value of 1.
Hence, in theory, if we would have infinitely many initial state landmarks, the heuristic value for all goals would be 1, independent from which landmarks were already achieved by the observed actions.
In contrast, when we completely ignore initial state landmarks (i.e., $|s_0| = 0$) the heuristic value for all goals \textit{only} depends on which non initial state landmarks exist for each goal and how many of those were already achieved by the observed actions.
Consequently, in this case, the decision is \textit{only} based on information that is gained from the observation sequence.
In summary, the more initial state landmarks there are compared to the number of non initial state landmarks, the less the decision on which goal(s) are the most probable ones depends on information that can be gained from the observation sequence.
How strongly the heuristic value for a goal is biased by considering initial state landmarks depends on how many non initial state landmarks exist for this goal.
If all goals would have similar numbers of non initial state landmarks, considering initial state landmarks would actually not affect the ranking of goals based on the completion heuristic.
Nevertheless, this assumption does not hold in almost all cases in practice, and due to this, we analyze the impact of the size of $l_g$ onto the heuristic score in the following.

How the value of $|l_g|$ affects the completion heuristic, again for the two extreme cases $|l_g|\to 0$ and $|l_g|\to\infty$, is formalized by equations \ref{eq:xToZero} and \ref{eq:xToInifinity}.
Moreover, in this formalization the case of $|al_g|=0$ for all goals $g\in G$ is considered.
Hence, we analyze how the completion heuristic behaves when no landmarks were achieved yet by the observation sequence.
\begin{equation}
\label{eq:xToZero}
    \lim_{|l_g|\to 0} \frac{|s_0|}{|l_g|+|s_0|} = 1
\end{equation}
When we consider $|l_g|\to 0$, i.e., there exist no non initial state landmarks for goal $g$, the completion heuristic converges to the value of 1.
This actually means, although we have not observed any evidence that goal $g$ is the actual goal of the agent, we decide that goal $g$ is the actual goal of the agent.
Of course, in practice, it is not very likely that $|l_g|=0$.
Nevertheless, this shows that the smaller the size of $l_g$ is, the closer the initial heuristic value is to 1.
\begin{equation}
\label{eq:xToInifinity}
    \lim_{|l_g|\to\infty} \frac{|s_0|}{|l_g|+|s_0|} = 0
\end{equation}
In contrast, when we consider $|l_g|\to\infty$, the initial heuristic value of the completion heuristic converges to 0.
In practice, this case will also not happen.
However, this means that the larger $|l_g|$ is compared to $|s_0|$, the closer the initial heuristic value will get to 0.
In summary, this analysis shows very well that by \textit{not} ignoring initial state landmarks, the completion heuristic heavily favors goals for which $|l_g|$ is small compared to $|s_0|$ when no landmarks were observed yet.
In addition, also the slope of the increase in the heuristic value depends on the size of $|l_g|$.
The smaller $|l_g|$ is, the higher will be the slope of the heuristic value increase.
Hence, by \textit{not} ignoring initial state landmarks, goals with small $|l_g|$ are not only heavily favored initially, but they also have a faster increase of heuristic values when non initial state landmarks are observed.

\section{Evaluation}
\label{sec:evaluation}
To evaluate the performance and efficiency of the adjusted methods discussed in the previous sections, we conducted several empirical experiments on three new benchmark datasets, which are based on a commonly used publicly available dataset \cite{zenodo}.
%To evaluate the performance of the landmark based hybrid goal recognition method and the amount of computation time required by the PLR approach compared to the RG method that is used in the state-of-the-art approach, we conducted an empirical experiment on a real-world data set (i.e., CMU Grand Kitchen Challenge \footnote{\url{http://kitchen.cs.cmu.edu/index.php}}).
More precisely, the goals of the evaluation are:
\begin{itemize}
    \item Show that ignoring initial state landmarks during the goal recognition process improves the recognition performance.
    \item Investigate how the structure of the benchmark problems affects goal recognition performance.
\end{itemize}

\subsection{Experimental Design}
\label{subsec:experimentalDesign}
%All experiments in this evaluation were carried out on machines that have 24 cores with 2.60GHz and at least 386GB RAM.
To assess the goal recognition performance of the different methods, we used the mean goal recognition precision.
We calculate the precision similar to existing works (e.g., \cite{amado2023robust})%, and consider a goal to be recognized correctly if it is part of the set of goals that were assigned with the highest heuristic score (i.e., $g_* \in \mathbf{G}$).
Furthermore, as we consider online goal recognition problems in this evaluation, we calculated the mean precision for different fractions $\lambda$ of total observations that were used for goal recognition.
Here, we used relative numbers because the lengths of the involved observation sequences substantially differ.
Hence, the mean precision Precision for a fraction $\lambda \in [0, 1]$ is calculated as follows:
\begin{equation}
\label{eq:Precision}
    Precision(\lambda,\mathcal{D}) = \frac{\sum_{R \in \mathcal{D}}{\frac{[g_{*R} \in R(\lfloor T_{R}\lambda\rfloor)]}{|\mathbf{G}_{R(T_R \lambda)}|}}}{|\mathcal{D}|}
\end{equation}
Here, $\mathcal{D}$ is a set of online goal recognition problems $R$, $g_{*R}$ denotes the correct goal of goal recognition problem $R$, $T_R$ is the maximum value of $t$ for online goal recognition problem $R$ (i.e., length of observation sequence that is associated with $R$), and $[g_{*R} \in R(t)]$ equals 1 if $g_{*R} \in \mathbf{G}_{R(t)}$ and 0 otherwise, where $\mathbf{G}_{R(t)}$ is the set of recognized goals for $R(t)$.
In other words, the precision quantifies the probability of picking the correct goal from a predicted set of goals $\mathbf{G}$ by chance.

\textit{Datasets.}
As the basis for our evaluation datasets, we use a dataset that is commonly used in the literature \cite{zenodo}.
However, we recognized that this dataset, which contains goal recognition problems from 13 different planning domains, almost only contains goal recognition problems that include only goals with similar size (i.e., the same number of facts in all possible goals).
Not only that this is not a very realistic scenario, as in practice one should expect that the different possible goals do not all have the same size in terms of facts they include.
In addition, this also biases the recognition performance of the original PLR approach, as in this case, the $l_g$ sets are more likely to have similar sizes.
To address this issue, we have created three new datasets that are based on the original dataset.
First, we modified the sets of possible goals in the existing dataset so that they have varying sizes.
During this process, we ensured that none of the possible goals is a true subgoal of any of the other possible goals in the same goal recognition problem.
Based on these modified goals, we created one dataset that has a random choice of true goals ($D_R$), one dataset in which the longest possible goals are the actual agent goals ($D_L$), and one dataset in which the shortest possible goals are the actual agent goals ($D_S$).
As we have discussed earlier, the original PLR approach does heavily favor goals with small $|l_g|$, which is more likely for goals that are smaller in general.
Hence, the original PLR approach should have an advantage in the third dataset.
To generate the observation sequences for the modified goals, we used the current version of the Fast Downward planner \cite{helmert2006fast}.

\subsection{Results}
\label{subsec:results}
\begin{figure}[h!]
	\centering
		\subfloat{
			\begin{tikzpicture}
			\pgfplotsset{every x tick label/.append style={font=\tiny}}
			\pgfplotsset{every y tick label/.append style={font=\tiny}}
				\begin{axis}[
					width=0.5\linewidth,
					height=5cm,
					ylabel={\scriptsize{Precision}},
					xlabel near ticks,
					ylabel near ticks,
					xmin=1, xmax=10,
					ymin=0, ymax=1.1,
					xtick={0,1,2,3,4,5,6,7,8,9,10, 11},
					xticklabels={0.0, 0.1, 0.2, 0.3, 0.4, 0.5, 0.6, 0.7, 0.8, 0.9, 1.0},
					ytick={0.2,0.4,0.6,0.8,1},
					legend pos=north west,
					ymajorgrids=false,
					xmajorgrids=false,
					major grid style={line width=.1pt,draw=gray!50},
					x axis line style={draw=black!60},
					tick style={draw=black!60},
					legend columns=4,
					legend style={draw=none},
					legend entries={\footnotesize{EX},\footnotesize{EX-init},\footnotesize{RHW},\footnotesize{RHW-init}},
					legend to name={average_performance}
				]

				\addplot[
					color=cyan,
						mark=none,
						dashed,
						]
						coordinates { (1.0,0.3528571428571428)(2.0,0.4078571428571429)(3.0,0.4707142857142858)(4.0,0.5007142857142858)(5.0,0.5807142857142857)(6.0,0.6528571428571429)(7.0,0.735)(8.0,0.7999999999999999)(9.0,0.8549999999999999)(10.0,0.9314285714285714)};

				\addplot[
					color=red,
						mark=none,
						dashed,
						]
						coordinates { (1.0,0.31285714285714283)(2.0,0.33285714285714285)(3.0,0.3935714285714286)(4.0,0.4278571428571429)(5.0,0.5007142857142857)(6.0,0.5564285714285714)(7.0,0.6592857142857144)(8.0,0.7371428571428572)(9.0,0.8192857142857142)(10.0,0.9235714285714286)};

				\addplot[
					color=orange,
						mark=none,
						dashed,
						]
						coordinates { (1.0,0.40071428571428575)(2.0,0.44)(3.0,0.5085714285714287)(4.0,0.5671428571428571)(5.0,0.6392857142857142)(6.0,0.7092857142857143)(7.0,0.7664285714285715)(8.0,0.8035714285714286)(9.0,0.8771428571428572)(10.0,0.9314285714285714)};

				\addplot[
					color=black,
						mark=none,
						dashed,
						]
						coordinates { (1.0,0.3514285714285715)(2.0,0.38714285714285707)(3.0,0.46571428571428575)(4.0,0.5285714285714286)(5.0,0.5942857142857143)(6.0,0.6742857142857144)(7.0,0.7335714285714285)(8.0,0.7728571428571429)(9.0,0.8614285714285715)(10.0,0.9235714285714286)};

				\end{axis}
			\end{tikzpicture}
			}
        \subfloat{
			\begin{tikzpicture}
			\pgfplotsset{every x tick label/.append style={font=\tiny}}
			\pgfplotsset{every y tick label/.append style={font=\tiny}}
				\begin{axis}[
					width=0.5\linewidth,
					height=5cm,
					xlabel near ticks,
					ylabel near ticks,
					xmin=1, xmax=10,
					ymin=0, ymax=1.1,
					xtick={0,1,2,3,4,5,6,7,8,9,10, 11},
					xticklabels={0.0, 0.1, 0.2, 0.3, 0.4, 0.5, 0.6, 0.7, 0.8, 0.9, 1.0},
					ytick={0.2,0.4,0.6,0.8,1},
					legend pos=north west,
					ymajorgrids=false,
					xmajorgrids=false,
					major grid style={line width=.1pt,draw=gray!50},
					x axis line style={draw=black!60},
					tick style={draw=black!60},
					legend columns=4,
					legend style={draw=none},
					legend entries={\footnotesize{EX},\footnotesize{EX-init},\footnotesize{RHW},\footnotesize{RHW-init}},
					legend to name={average_performance}
				]

				\addplot[
					color=cyan,
						mark=none,
						dashed,
						]
						coordinates { (1.0,0.27714285714285714)(2.0,0.31142857142857144)(3.0,0.3450000000000001)(4.0,0.37071428571428566)(5.0,0.42642857142857143)(6.0,0.4407142857142857)(7.0,0.49428571428571433)(8.0,0.5414285714285715)(9.0,0.5842857142857143)(10.0,0.6285714285714284)};

				\addplot[
					color=red,
						mark=none,
						dashed,
						]
						coordinates { (1.0,0.20285714285714287)(2.0,0.2292857142857143)(3.0,0.24499999999999997)(4.0,0.2564285714285714)(5.0,0.2664285714285714)(6.0,0.28428571428571425)(7.0,0.29071428571428576)(8.0,0.3007142857142857)(9.0,0.315)(10.0,0.3242857142857143)};

				\addplot[
					color=orange,
						mark=none,
						dashed,
						]
						coordinates { (1.0,0.26999999999999996)(2.0,0.3242857142857143)(3.0,0.3392857142857143)(4.0,0.3657142857142857)(5.0,0.39428571428571424)(6.0,0.4049999999999999)(7.0,0.42714285714285716)(8.0,0.4364285714285714)(9.0,0.4742857142857143)(10.0,0.4785714285714286)};

				\addplot[
					color=black,
						mark=none,
						dashed,
						]
						coordinates { (1.0,0.20357142857142857)(2.0,0.2307142857142857)(3.0,0.2535714285714285)(4.0,0.24499999999999997)(5.0,0.27714285714285714)(6.0,0.2835714285714286)(7.0,0.2885714285714286)(8.0,0.30642857142857144)(9.0,0.3142857142857142)(10.0,0.3221428571428571)};

				\end{axis}
			\end{tikzpicture}
			}
   \newline
   \subfloat{
			\begin{tikzpicture}
			\pgfplotsset{every x tick label/.append style={font=\tiny}}
			\pgfplotsset{every y tick label/.append style={font=\tiny}}
				\begin{axis}[
					width=0.5\linewidth,
					height=5cm,
					ylabel={\scriptsize{Precision}},
					xlabel near ticks,
					ylabel near ticks,
					xmin=1, xmax=10,
					ymin=0, ymax=1.1,
					xtick={0,1,2,3,4,5,6,7,8,9,10, 11},
					xticklabels={0.0, 0.1, 0.2, 0.3, 0.4, 0.5, 0.6, 0.7, 0.8, 0.9, 1.0},
					ytick={0.2,0.4,0.6,0.8,1},
					legend pos=north west,
					ymajorgrids=false,
					xmajorgrids=false,
					major grid style={line width=.1pt,draw=gray!50},
					x axis line style={draw=black!60},
					tick style={draw=black!60},
					legend columns=4,
					legend style={draw=none},
					legend entries={\footnotesize{EX},\footnotesize{EX-init},\footnotesize{RHW},\footnotesize{RHW-init}},
					legend to name={average_performance}
				]

				\addplot[
					color=cyan,
						mark=none,
						dashed,
						]
						coordinates { (1.0,0.25)(2.0,0.32642857142857146)(3.0,0.3992857142857143)(4.0,0.4457142857142858)(5.0,0.527142857142857)(6.0,0.5928571428571429)(7.0,0.6685714285714285)(8.0,0.7321428571428571)(9.0,0.812142857142857)(10.0,0.9)};

				\addplot[
					color=red,
						mark=none,
						dashed,
						]
						coordinates { (1.0,0.1542857142857143)(2.0,0.17785714285714288)(3.0,0.24571428571428575)(4.0,0.2657142857142857)(5.0,0.3442857142857143)(6.0,0.3985714285714286)(7.0,0.5035714285714286)(8.0,0.6228571428571429)(9.0,0.7514285714285716)(10.0,0.8928571428571429)};

				\addplot[
					color=orange,
						mark=none,
						dashed,
						]
						coordinates { (1.0,0.2521428571428571)(2.0,0.31857142857142856)(3.0,0.4035714285714285)(4.0,0.4557142857142858)(5.0,0.5514285714285714)(6.0,0.6207142857142858)(7.0,0.6950000000000001)(8.0,0.7571428571428571)(9.0,0.8457142857142858)(10.0,0.9)};

				\addplot[
					color=black,
						mark=none,
						dashed,
						]
						coordinates { (1.0,0.22071428571428572)(2.0,0.26071428571428573)(3.0,0.34714285714285714)(4.0,0.4028571428571429)(5.0,0.49142857142857144)(6.0,0.5721428571428572)(7.0,0.6449999999999999)(8.0,0.7071428571428572)(9.0,0.8192857142857143)(10.0,0.8935714285714286)};

				\end{axis}
			\end{tikzpicture}
			}
   \subfloat{
			\begin{tikzpicture}
			\pgfplotsset{every x tick label/.append style={font=\tiny}}
			\pgfplotsset{every y tick label/.append style={font=\tiny}}
				\begin{axis}[
					width=0.5\linewidth,
					height=5cm,
					xlabel near ticks,
					ylabel near ticks,
					xmin=1, xmax=10,
					ymin=0, ymax=1.1,
					xtick={0,1,2,3,4,5,6,7,8,9,10, 11},
					xticklabels={0.0, 0.1, 0.2, 0.3, 0.4, 0.5, 0.6, 0.7, 0.8, 0.9, 1.0},
					ytick={0.2,0.4,0.6,0.8,1},
					legend pos=north west,
					ymajorgrids=false,
					xmajorgrids=false,
					major grid style={line width=.1pt,draw=gray!50},
					x axis line style={draw=black!60},
					tick style={draw=black!60},
					legend columns=4,
					legend style={draw=none},
					legend entries={\footnotesize{EX},\footnotesize{EX-init},\footnotesize{RHW},\footnotesize{RHW-init}},
					legend to name={average_performance}
				]

				\addplot[
					color=cyan,
						mark=none,
						dashed,
						]
						coordinates { (1.0,0.3321428571428572)(2.0,0.4042857142857143)(3.0,0.45285714285714285)(4.0,0.49214285714285705)(5.0,0.5564285714285714)(6.0,0.5785714285714285)(7.0,0.6264285714285714)(8.0,0.6628571428571428)(9.0,0.6964285714285714)(10.0,0.7221428571428572)};

				\addplot[
					color=red,
						mark=none,
						dashed,
						]
						coordinates { (1.0,0.3257142857142857)(2.0,0.36357142857142855)(3.0,0.37714285714285717)(4.0,0.3871428571428571)(5.0,0.4164285714285714)(6.0,0.4321428571428572)(7.0,0.45214285714285707)(8.0,0.4707142857142857)(9.0,0.4885714285714285)(10.0,0.49571428571428566)};

				\addplot[
					color=orange,
						mark=none,
						dashed,
						]
						coordinates { (1.0,0.3885714285714285)(2.0,0.4171428571428572)(3.0,0.47428571428571425)(4.0,0.49999999999999994)(5.0,0.5378571428571428)(6.0,0.5635714285714285)(7.0,0.5842857142857143)(8.0,0.5935714285714286)(9.0,0.6364285714285715)(10.0,0.6535714285714286)};

				\addplot[
					color=black,
						mark=none,
						dashed,
						]
						coordinates { (1.0,0.32142857142857145)(2.0,0.37214285714285705)(3.0,0.3921428571428572)(4.0,0.41500000000000004)(5.0,0.4342857142857143)(6.0,0.45000000000000007)(7.0,0.46142857142857135)(8.0,0.48428571428571426)(9.0,0.4928571428571428)(10.0,0.5207142857142857)};

				\end{axis}
			\end{tikzpicture}
			}
   \newline
   \subfloat{
			\begin{tikzpicture}
			\pgfplotsset{every x tick label/.append style={font=\tiny}}
			\pgfplotsset{every y tick label/.append style={font=\tiny}}
				\begin{axis}[
					width=0.5\linewidth,
					height=5cm,
					xlabel={\scriptsize{$\lambda$}},
					ylabel={\scriptsize{Precision}},
					xlabel near ticks,
					ylabel near ticks,
					xmin=1, xmax=10,
					ymin=0, ymax=1.1,
					xtick={0,1,2,3,4,5,6,7,8,9,10, 11},
					xticklabels={0.0, 0.1, 0.2, 0.3, 0.4, 0.5, 0.6, 0.7, 0.8, 0.9, 1.0},
					ytick={0.2,0.4,0.6,0.8,1},
					legend pos=north west,
					ymajorgrids=false,
					xmajorgrids=false,
					major grid style={line width=.1pt,draw=gray!50},
					x axis line style={draw=black!60},
					tick style={draw=black!60},
					legend columns=4,
					legend style={draw=none},
					legend entries={\footnotesize{EX},\footnotesize{EX-init},\footnotesize{RHW},\footnotesize{RHW-init}},
					legend to name={average_performance}
				]

				\addplot[
					color=cyan,
						mark=none,
						dashed,
						]
						coordinates { (1.0,0.37785714285714284)(2.0,0.44071428571428567)(3.0,0.495)(4.0,0.5307142857142857)(5.0,0.6192857142857143)(6.0,0.6742857142857142)(7.0,0.7328571428571429)(8.0,0.8064285714285714)(9.0,0.8685714285714287)(10.0,0.9357142857142858)};

				\addplot[
					color=red,
						mark=none,
						dashed,
						]
						coordinates { (1.0,0.35500000000000004)(2.0,0.3885714285714285)(3.0,0.4457142857142857)(4.0,0.4714285714285714)(5.0,0.5428571428571429)(6.0,0.6178571428571428)(7.0,0.6900000000000002)(8.0,0.7692857142857142)(9.0,0.8442857142857144)(10.0,0.9292857142857142)};

				\addplot[
					color=orange,
						mark=none,
						dashed,
						]
						coordinates { (1.0,0.4585714285714286)(2.0,0.485)(3.0,0.5442857142857143)(4.0,0.5828571428571429)(5.0,0.6807142857142857)(6.0,0.7335714285714287)(7.0,0.7764285714285715)(8.0,0.8192857142857142)(9.0,0.8857142857142859)(10.0,0.9357142857142858)};

				\addplot[
					color=black,
						mark=none,
						dashed,
						]
						coordinates { (1.0,0.40285714285714286)(2.0,0.44571428571428573)(3.0,0.5135714285714286)(4.0,0.5592857142857144)(5.0,0.6421428571428572)(6.0,0.7114285714285715)(7.0,0.7585714285714287)(8.0,0.7964285714285716)(9.0,0.8842857142857145)(10.0,0.9292857142857142)};

				\end{axis}
			\end{tikzpicture}
			}
   \subfloat{
			\begin{tikzpicture}
			\pgfplotsset{every x tick label/.append style={font=\tiny}}
			\pgfplotsset{every y tick label/.append style={font=\tiny}}
				\begin{axis}[
					width=0.5\linewidth,
					height=5cm,
					xlabel={\scriptsize{$\lambda$}},
					xlabel near ticks,
					ylabel near ticks,
					xmin=1, xmax=10,
					ymin=0, ymax=1.1,
					xtick={0,1,2,3,4,5,6,7,8,9,10, 11},
					xticklabels={0.0, 0.1, 0.2, 0.3, 0.4, 0.5, 0.6, 0.7, 0.8, 0.9, 1.0},
					ytick={0.2,0.4,0.6,0.8,1},
					legend pos=north west,
					ymajorgrids=false,
					xmajorgrids=false,
					major grid style={line width=.1pt,draw=gray!50},
					x axis line style={draw=black!60},
					tick style={draw=black!60},
					legend columns=4,
					legend style={draw=none},
					legend entries={\footnotesize{EX},\footnotesize{EX-init},\footnotesize{RHW},\footnotesize{RHW-init}},
					legend to name={average_performance}
				]

				\addplot[
					color=cyan,
						mark=none,
						dashed,
						]
						coordinates { (1.0,0.24785714285714286)(2.0,0.2814285714285714)(3.0,0.30785714285714283)(4.0,0.31714285714285717)(5.0,0.3542857142857142)(6.0,0.41)(7.0,0.42428571428571427)(8.0,0.4928571428571429)(9.0,0.5542857142857143)(10.0,0.5935714285714285)};

				\addplot[
					color=red,
						mark=none,
						dashed,
						]
						coordinates { (1.0,0.14857142857142858)(2.0,0.17142857142857146)(3.0,0.17357142857142857)(4.0,0.16999999999999998)(5.0,0.19)(6.0,0.1978571428571429)(7.0,0.2035714285714286)(8.0,0.2178571428571429)(9.0,0.22571428571428573)(10.0,0.23857142857142857)};

				\addplot[
					color=orange,
						mark=none,
						dashed,
						]
						coordinates { (1.0,0.24571428571428572)(2.0,0.275)(3.0,0.2828571428571429)(4.0,0.2992857142857143)(5.0,0.32428571428571434)(6.0,0.3478571428571429)(7.0,0.3685714285714286)(8.0,0.3771428571428572)(9.0,0.4035714285714285)(10.0,0.4278571428571428)};

				\addplot[
					color=black,
						mark=none,
						dashed,
						]
						coordinates { (1.0,0.14142857142857143)(2.0,0.16071428571428573)(3.0,0.1592857142857143)(4.0,0.16357142857142856)(5.0,0.19571428571428573)(6.0,0.19785714285714287)(7.0,0.2121428571428572)(8.0,0.22142857142857145)(9.0,0.22571428571428573)(10.0,0.2335714285714286)};

				\end{axis}
			\end{tikzpicture}
			}
			\newline
			\vspace{-0.2cm}
	\ref{average_performance}
	\caption{Average precision of the Exhaustive (EX), Exhaustive with initial state landmarks (EX-init), Richter (RHW), and Richter with initial state landmarks (RHW-init) approaches on the three benchmark datasets $D_R$, $D_L$, and $D_S$ (depictured in this order from top to bottom). The three subfigures on the left handside report the results for the completion heuristic, while the subfigures on the right handside report the results for the uniqueness heuristic.}
	\label{fig:average_performance}
\end{figure}
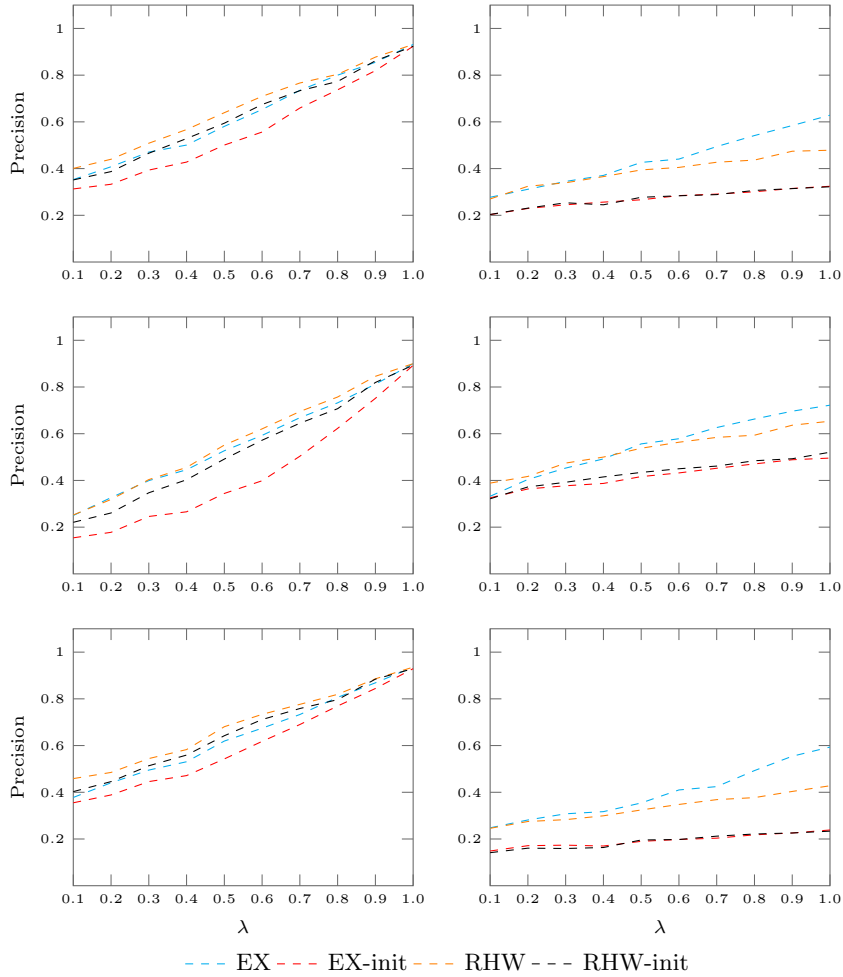
Figure \ref{fig:average_performance} shows the average precision for $D_R$, $D_L$, and $D_S$ (depicted in this order from top to bottom) over all 13 domains in each benchmark dataset.
On the left hand side, the average precision for the completion heuristic is reported and on the right hand side, the average precision for the uniqueness heuristic is depicted.
Further, the subfigure for each combination of heuristic and dataset shows the average precision for each of the evaluated approaches (Exhaustive (EX), Exhaustive with initial state landmarks (EX-init), Richter (RHW), and Richter with initial state landmarks (RHW-init)).
The results show that ignoring initial state landmarks for goal recognition, as we propose in this paper, leads for each evaluated combination of approach, dataset, and heuristic to superior recognition performance compared to when initial state landmarks \textit{are used} for recognition.
Interesting to note is also that the performance difference is larger for the EX extraction algorithm than for the RHW extraction algorithm.
One reason for this might be that the RHW algorithm allows for disjunctive landmarks, whereas the EX algorithm only considers single fact landmarks.

As expected, the completion heuristic achieves the best results on the $D_S$ dataset, in which the shortest goals are the true goals in the goal recognition setups.
This is, as analyzed previously, due to the way in which the completion heuristic favors goals with smaller sets of landmarks.
Due to the same reason, the results show that the performance difference between ignoring initial state landmarks and not ignoring them is the largest for the $D_L$ dataset.
In contrast, interestingly, the uniqueness heuristic seems to favor goals with larger sets of landmarks.
Most probably this is due to the weighting through the uniqueness scores that this heuristic uses.
It is very likely that goals with larger landmark sets also have more facts in their goal description than goals with smaller landmark sets.
As the agent always starts at the same initial state, the shorter the plan (which in many domains correlates with goal description size) the more likely it becomes that the goal of this plan shares landmarks with other goals and hence, has less unique landmarks.
Respectively, the longer a plan becomes, the more likely it is that the goal of such a plan includes more unique landmarks in its set of landmarks.
Consequently, this leads to the uniqueness heuristic favoring longer goals.
%Also interesting to note is that the recognition performance difference between ignoring initial state landmarks and not ignoring them is even more prominent for the uniqueness heuristic than for the completion heuristic.

\section{Related Work}
\label{sec:relatedwork}
Since the idea of Plan Recognition as Planning was introduced by \cite{ramirez2009plan}, many approaches have adopted this paradigm \cite{ramirez2010probabilistic}, \cite{ramirez2011goal}, \cite{yolanda2015fast}, \cite{vered2016mirroring}, \cite{sohrabi2016revisited}, \cite{masters2017cost}, \cite{pereira2020landmark}, \cite{cohausz2022plan}.
It was recognized relatively soon that the initial PRAP approaches are computationally demanding, as they require computing entire plans.
Since then, this problem has been addressed by many studies with the approach by \cite{pereira2020landmark} being a recent example.
This method also belongs to a recent type of PRAP methods (to which VS belongs as well), which do not derive probability distributions over the set of possible goals by analyzing cost differences but rank the possible goals by calculating heuristic values.
Another approach from this area is a variant that was suggested as an approximation for their main approach by \cite{ramirez2009plan}.

\section{Conclusion}
\label{sec:conclusion}
In conclusion, in this paper we have formally analyzed and discussed why using initial state landmarks for goal recognition biases the recognition performance.
Moreover, we provided three new benchmark datasets, which are based on a dataset that is commonly used in the literature \cite{zenodo}..
These three benchmark datasets were used to empirically show that ignoring initial state landmarks for goal recognition is indeed superior regarding goal recognition performance.
In addition, we empirically evaluated the effect of different goal recognition problem structures onto the goal recognition performance of planning landmark based goal recognition approaches.
An interesting avenue for future work would to evaluate how well the slightly adjusted algorithms, proposed in this paper, handle missing and/or noisy observations.

\
 \bibliographystyle{splncs04}
 \bibliography{ki23}
\end{document}